\documentclass{article}

\usepackage{enumitem}
\usepackage{amsmath}
\usepackage{multirow} 
\usepackage{booktabs}
\usepackage[utf8]{inputenc}
\usepackage{algorithm}
\usepackage{algpseudocode}
\usepackage{graphicx}
\usepackage{tabularx}
\usepackage[preprint]{neurips_2024}

\usepackage[utf8]{inputenc} % allow utf-8 input
\usepackage[T1]{fontenc}    % use 8-bit T1 fonts
\usepackage{hyperref}       % hyperlinks
\usepackage{url}            % simple URL typesetting
\usepackage{booktabs}       % professional-quality tables
\usepackage{amsfonts}       % blackboard math symbols
\usepackage{nicefrac}       % compact symbols for 1/2, etc.
\usepackage{microtype}      % microtypography
\usepackage{xcolor}         % colors

\title{LangTopo: Aligning \textbf{Lang}uage Descriptions of Graphs with Tokenized \textbf{Topo}logical Modeling}

\author{%
  Zhong Guan\\
    Tianjing University
  \And
  Hongke Zhao \\
  Tianjing University \\
   \AND
   Likang Wu \\
   University of Science and Technology of China \\
   \And
    Ming He \\
    AI Lab at Lenovo Research \\
    \And
    Jianpin Fan \\
    AI Lab at Lenovo Research \\
}

\begin{document}

\maketitle

\begin{abstract}
Recently, large language models (LLMs) have been widely researched in the field of graph machine learning due to their outstanding abilities in language comprehension and learning. However, the significant gap between natural language tasks and topological structure modeling poses a nonnegligible challenge. Specifically, since natural language descriptions are not sufficient for LLMs to understand and process graph-structured data, fine-tuned LLMs perform even worse than some traditional GNN models on graph tasks, lacking inherent modeling capabilities for graph structures. Existing research overly emphasizes LLMs' understanding of semantic information captured by external models, while inadequately exploring graph topological structure modeling, thereby overlooking the genuine capabilities that LLMs lack. Consequently, in this paper, we introduce a new framework, LangTopo, which aligns graph structure modeling with natural language understanding at the token level. LangTopo quantifies the graph structure modeling capabilities of GNNs and LLMs by constructing a codebook for the graph modality and performs consistency maximization. This process aligns the text description of LLM with the topological modeling of GNN, allowing LLM to learn the ability of GNN to capture graph structures, enabling LLM to handle graph-structured data independently. We demonstrate the effectiveness of our proposed method on multiple datasets.
\end{abstract}

\section{INTRODUCTION}

Text-attributed graphs (TAGs) are ubiquitous in the real world~\cite{berge2001theory}, appearing in academic citation networks~\cite{wang2020microsoft}, web pages~\cite{mernyei2020wiki}, e-commerce platforms~\cite{shchur2018pitfalls}, and search services. These graphs are characterized by nodes that contain rich textual attributes, making them complex and challenging to analyze. Research on TAGs has become a significant area within graph machine learning, primarily focusing on Graph Neural Networks (GNNs) based on message-passing mechanisms to exploit adjacency spaces effectively~\cite{defferrard2016convolutional,kipf2016semi,velivckovic2017graph}.

Recently, the advent of large language models (LLMs) like ChatGPT~\cite{openai2023gpt} and Llama~\cite{touvron2023llama} has sparked significant interest in their application to graph machine learning tasks due to their impressive potential across various fields~\cite{23,24}. Some studies have explored enhancing node embeddings~\cite{duan2023simteg,he2023harnessing} and reinforcing graph topological structures using LLMs~\cite{guo2024graphedit,wang2024llm}. However, unlike GNNs, which are based on message passing, LLMs grounded in natural language often struggle to describe and process node connections, leading to suboptimal performance on graph-related tasks~\cite{chen2024exploring,huang2023can}.

\begin{figure}
\centering
\includegraphics[width=1\linewidth]{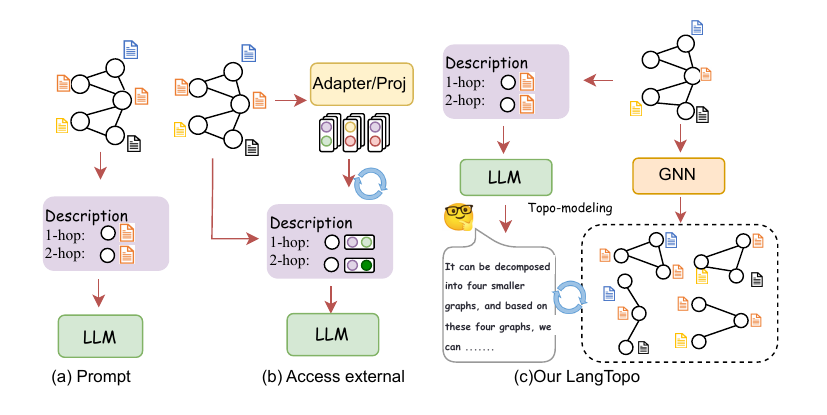}
\caption{\textbf{Prompt:} LLMs make predictions based solely on natural language descriptions. \textbf{Access external:} LLMs leverage external models (typically GNNs) to extract information for enhanced predictions. \textbf{Ours LangTopo:} Aligning the textual descriptive power of LLMs with the topological modeling capabilities of GNNs in terms of model processing and operation.}
\label{fig:intro}
\end{figure}

Combining the structural modeling capacity of GNNs with the text processing capability of LLMs presents a promising approach to addressing these challenges. A straightforward solution involves using an external GNN to extract spatial structure embeddings, followed by training a projection layer or adapter to inject these embeddings into the LLM, as illustrated in Figure~\ref{fig:intro}(b). Recent approaches have enhanced LLM performance in graph tasks through such methods~\cite{chen2024llaga,huang2024can,zhang2024graphtranslator}. However, LLMs still lack the ability to handle graph data independently and continue to rely on external models during inference. Furthermore, aligning the embedding spaces is redundant in the context of text-attributed graphs because GNNs' embedding spaces already encapsulate natural language information. The fundamental issue is that LLMs lack the capability to model graph structures. Consequently, we posit that the crux of the problem is modeling, not embedding.

To address this issue, we propose LangTopo, a novel learning paradigm that aligns language descriptions of graphs with tokenized topological modeling from a view of consistent modeling methods for text and structure. Initially, we employ a graph neural network to introduce a modality-specific codebook for graphs, capable of quantizing both textual and spatial information of nodes. By incorporating Gumbel-softmax~\cite{jang2016categorical}, we transform discrete quantization into continuous, differentiable quantization through relaxation techniques.
Through this process, we efficiently extract the modeling prowess of GNNs into two components: \textbf{Relaxed Distribution}, indicating the allocation patterns of subgraph structures relative to the codebook, and \textbf{Quantized Embeddings}, representing precise subgraph structures with respect to the codebook. Subsequently, the LLM describes the graph structure in natural language and obtains these two components based on the codebook embeddings. Driven by the goal of achieving similarity between the corresponding components produced by the LLM and GNN, LangTopo aligns the LLM’s natural language description of the graph structure with the graph structure modeled by GNN, ultimately enhancing the LLM’s aptitude for modeling spatial architectures.

The main contributions of this work are summarized below:

\begin{itemize}[leftmargin=*]
\item We propose LangTopo, a novel framework for learning graph structures using LLMs. Through the adoption of Vector Quantized-Variational Autoencoder (VQ-VAE), we quantify the modeling capabilities of LLMs and GNNs for graph topological structures and enable LLMs to learn GNNs' ability to model graph structures through supervised learning.
\item We achieve alignment between the natural language descriptive text in LLMs and the processing and operation of GNN models by constructing a codebook for the graph data modality and thereby ensuring that the quantized embeddings and relaxation distributions are similar between LLMs and GNNs.
\item Unlike existing paradigms that usually introduce external modules to recognize graph structures, LangTopo endows the LLM itself with the ability to model graph structures, obviating the need for external data or model integration during inference. LangTopo demonstrates excellent results across multiple datasets.
\end{itemize}

\section{RELARED WORK}
\textbf{LLM for Graph.   } Recent studies have explored the application of Large Language Models (LLMs) in the field of graph structures~\cite{wu2024exploring}. GLEM~\cite{zhao2022learning} and others~\cite{yang2021graphformers,li2021adsgnn} have investigated the joint training of LLMs and graph neural networks (GNNs). TAPE~\cite{he2023harnessing} utilizes LLMs to predict the ranking classification of nodes and provides detailed explanations to enhance the embedding quality of GNNs. Sun et al.\cite{sun2023large} leverage LLMs for pseudo-labeling generation to improve the topological structure information of graphs. Furthermore, more research focuses on enhancing LLMs’ direct processing of text graphs. InstructGLM~\cite{ye2023natural} first uses LLM-based instruction tuning to describe the structure of graphs and node features, and solve graph tasks. GraphText~\cite{zhao2023graphtext} introduced a generic graph reasoning framework for in-context learning and directed fine-tuning of graph structures. Despite these advancements, LLMs struggle with structured data translated into natural language, often yielding suboptimal results. To overcome this challenge, LLaGA~\cite{chen2024llaga} reformulates node link information as sequential data, applying instruction tuning that LLMs can comprehend while maintaining structural node information. UniGraph~\cite{he2024unigraph} adopts a masked strategy for joint LLM and GNN training under co-training, achieving generalization across various graphs and datasets. GraphAdapter~\cite{huang2024can} employs a GNN model as an adapter, collaborates with LLMs for TAG tasks, and facilitates task-specific fine-tuning through external access. Nevertheless, these methods do not empower LLMs with the ability to improve themselves, and LLMs still lack the knowledge and capabilities to handle graph data structures.

\textbf{Graph neural network.   }  Graph Neural Networks (GNNs), standing at the forefront of graph machine learning, have solidified their position as paramount instruments in graph-based learning scenarios. Their prowess lies in effectively modeling complex relational structures within graphs. Pioneering work was conducted by Kipf et al~\cite{kipf2016semi}. with the development of Graph Convolutional Networks (GCNs), ingeniously aggregating features from one-hop neighbors for each node. Following suit, the advent of GAT~\cite{velivckovic2017graph} and GraphSAGE~\cite{defferrard2016convolutional} marked significant success in tackling graph learning challenges, with a common strategy of using a message passing mechanism. Based on this, research has further ventured into addressing issues of heterophily~\cite{luan2022revisiting,zhu2021graph,li2022finding,ma2021homophily,wu2023learning} in graph learning tasks and the oversmoothing phenomenon in graph convolutions~\cite{li2021training,liu2020towards,chen2020measuring,giraldo2023trade}.

\section{PRELIMINARIES}
\textbf{Vector Quantised-Variational AutoEncoder (VQ-VAE).  } VQ-VAE~\cite{van2017neural} is a vector quantizer that generates discrete codes by applying quantization to encoded continuous variables and is capable of reconstructing the original data based on the latent variables corresponding to the discrete codes. It is trained jointly by updating the quantization codebook as well as the encoder and decoder. Formally, VQ-VAE first encodes the input $x$ via encoder into  $z_e(x)$. Then, based on the nearest neighbor lookup between $z_e(x)$ and the codebook $e_i$, $i=1,2,…,K$, the potential variable $z_q(x)$is found ($z_q(x)=e_i$, where $i= argmin_j ||z_e(x)-e_j||$). The potential variable $z_q(x)$ is then passed to the decoder, which attempts to reconstruct the input $x$. VQ-VAE, like the VAE, optimizes the Evidence Lower Bound (ELBO)~\cite{kingma2013auto}. 
\begin{equation}
\label{eq:eq1}
\mathcal{L}(\theta, \phi; x) = -D_{KL}(q_{\phi}(z|x) || p_{\theta}(z)) + \mathbb{E}_{q_{\phi}(z|x)}[\log p_{\theta}(x|z)]
\end{equation}
 At the same time, some modifications were made, and the loss function is as follows:
\begin{equation}
\label{eq:eq2}
\mathcal{L}(\mathbf{x}) = \mathcal{L}_{\text{recon}} + ||\text{sg}[\mathbf{z}_e(\mathbf{x})] - \mathbf{z}_q(\mathbf{x})||_2^2 + \beta \cdot ||\text{sg}[\mathbf{z}_q(\mathbf{x})] - \mathbf{z}_e(\mathbf{x})||_2^2
\end{equation}
wherein, $\mathcal{L}_{\text{recon}}$is the reconstruction loss function, $||\text{sg}[\mathbf{z}_e(\mathbf{x})] - \mathbf{z}_q(\mathbf{x})||_2^2$is the codebook loss, used only for updating the codebook to make the selected  $\mathbf{z}_q$ close to the encoder’s output $\mathbf{z}_e$, where sg is the stop-gradient operation. $\beta \cdot ||\text{sg}[\mathbf{z}_q(\mathbf{x})] - \mathbf{z}_e(\mathbf{x})||_2^2$is the commitment loss, applicable only to the encoder weights, which encourages the encoder’s output to stay close to the selected code, preventing it from fluctuating too frequently between code vectors. Additionally, since the prior distribution is uniform , and each codeword in $q_{\phi}(z|x) $ is a discrete integer, the KL divergence is constant.

\textbf{Text-Attributed Graph(TAG).   } We define a TAG as a graph where each node has corresponding text. Define TAG as $G = (V, A, X)$ with nodes $V$ and adjacency matrix $A \in \mathbb{R}^{|V| \times |V|}$, and node attributes as the $|V| \times d$ feature matrix $X$. In this paper, our research objective is node classification in TAG. Given a few labeled nodes $y_L$ of $L \subseteq V$, the goal is to predict the labels $y_U$ for the remaining unlabeled objects $U = V \setminus L$.

\begin{figure}[t]
    \centering
    \includegraphics[width=1\linewidth]{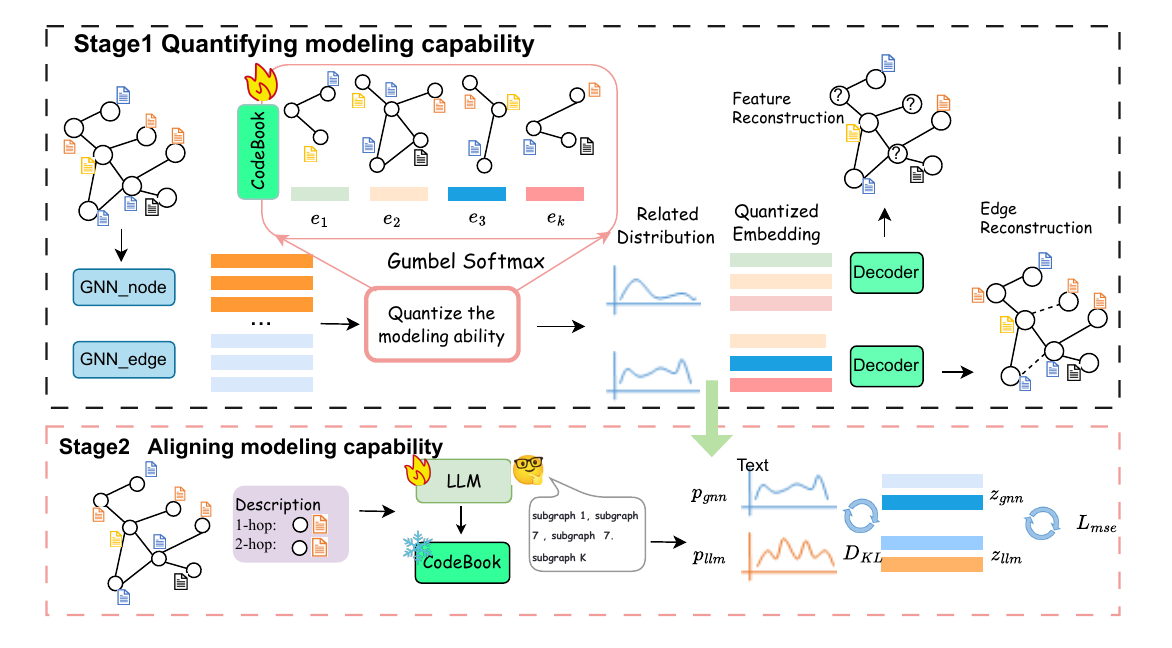}
    \caption{The model architecture of our proposed LangTopo framework for graph structure learning.}
    \label{fig:main}
\end{figure}

\section{METHODOLOGY}
Our framework consists of two parts:

(i)\textit{ Quantization of Modeling Ability.} By training the codebook based on the textual and spatial information of the TAG, we obtain quantized embeddings and relaxed distributions, alongside a codebook that encapsulates diverse graph structural information. These quantized embeddings and relaxed distributions represent the model’s capability to model graph structures.

(ii)\textit{ LLM’s graph structure learning.   }Train LLM adapted for graph tasks and evaluate its graph structural modeling ability using  established codebook. Further, improve the graph structural modeling capability through alignment with GNN’s quantized embeddings and relaxed distributions.

\subsection{Quantization of GNN Modeling Ability}
In this section, we will discuss the training of a codebook that accounts for both node text and spatial structure using VQ-VAE. Through the graph modality-specific codebook, we obtain quantized embeddings and relaxed distributions that represent the GNN’s ability to model graph structure. As illustrated in Stage 1 of Figure~\ref{fig:main}

\textbf{4.1.1 Quantization of Text and Spatial Structure.     } First, we extract node embeddings $h_{node}^{(l)}$ and $h_{edge}^{(l)}$ by multi-layer aggregation through two GNN models. Then, according to the specific quantization selection method, we obtain a codeword $z_{node}$, $z_{edge}$ from $E = {e_1, e_2,...,e_K}$. This can be described as follows: 
\begin{equation}
\label{eq:eq3}
h_{node}^{(l)}=\sigma(AGG_\phi(MSG_\phi(h_{{\text{NB}}(node)}^{(l-1)},A), 
h_{edge}^{(l)}=\sigma(AGG_\theta(MSG_\theta(h_{{\text{NB}}(edge)}^{(l-1)},A)
\end{equation}
\begin{equation}
\label{eq:eq4}
z_{node}=Lookup(E,h_{node}),
z_{edge}=Lookup(E,h_{edge})
\end{equation}
Here, $\theta$ and $\phi$ respectively denote the GNN models focusing on node text information and spatial structure, while "Lookup" represents the method for selecting discretized codewords from the codebook. In the context of VQ-VAE, the embedding of the codebook is obtained through a nearest neighbor search. To enhance the representativeness of the quantized vectors, Section 4.1.2 introduces the utilization of the Gumbel-softmax relaxation technique for acquiring codebook vectors.

Subsequently, the codebook vectors $z_{node}$ and $z_{edge}$ are respectively passed through a simple linear projection layer ($p_{\omega}:z->\hat{v_i}$) for node feature reconstruction and adjacency matrix reconstruction.

\textbf{4.1.2 Gumbel softmax related disturbtion.  }
Gumbel Softmax relaxation is a technique for handling discrete variables by allowing us to approximate the probability distribution of a discrete variable using a continuous, differentiable function. The discretization step in VQ-VAE, where a single codeword is selected, inherently neglects the structural information encapsulated by other codewords in the codebook, thereby limiting the comprehensive assessment of a GNN's capacity to model intricate graphs. This shortfall motivates our choice of employing the Gumbel-Softmax relaxation technique. Formally, it can be expressed as:
\begin{equation}
\label{eq:eq5}
g_i = -\log\left(-\log(u)\right),
u \sim Uniform(0,1)
\end{equation}
\begin{equation}
\label{eq:eq6}
p_i = \frac{\exp\left((\log(\pi_i) + g_i)/\tau\right)}{\sum_{j=1}^{K} \exp\left((\log(\pi_j) + g_j)\tau\right)} , \quad i=1,2,...,K
\end{equation}
\begin{equation}
\label{eq:eq7}
z = \sum_{j=1}^{k} {p_j}{z_j} ,  j=1,2,...,k
\end{equation}
Wherein, $g_i$ denotes a sample from the Gumbel distribution, $p_i$ represents  the probability of selecting each codeword, and $z$ signifies  the quantized embedding. Following the application of Gumbel-Softmax, we obtain the quantized embedding $Z$ and the relaxed distribution $P$, which effectively represent the model’s modeling of the graph modality. The temperature coefficient $\tau$ significantly governs the relaxation level of the distribution: higher values of $\tau$ promote a distribution closer to uniformity, whereas diminished values steer the distribution towards a sharper, nearly one-hot form. In our experiments, we adopt an annealing strategy to gradually decrease the value of $\tau$  over time.

We demonstrate in 
\textbf{Appendix} \ref{sec:appendixA}  that after using Gumbel softmax, the new random variable $z$ is the same as the random variable $\pi$, that is, the probability of taking $z$ is the same as the probability of taking $\pi$.

\textbf{4.1.3 Optimization.  }  
Similar to VAE, we also optimize the ELBO(Eq~\ref{eq:eq1}), with the loss function as follows:
\begin{equation}
\begin{aligned}
\label{eq:eq8}
\mathcal{L}_{Stage1}=\mathbb{E}_{q_{\phi}(z|x)}[\log p_{\theta}(x|z)]-D_{KL}(q_{\phi}(z|x) || p_{\theta}(z)) \\
=\mathcal{L}_{edge\_recon}+\alpha_{node} \mathcal{L}_{node\_recon}+\beta_{KL}  KL(q_{\phi}(z|x) || p_{\theta}(z))\\
\end{aligned}
\end{equation}
\begin{equation}
\label{eq:eq100}
\mathcal{L}_{node\_recon}=\underbrace{\frac{1}{N}\sum_{i=1}^{N}(1 - \frac{v_i^T\hat{v}_i}{\|v_i\|\cdot\|\hat{v}_i\|})^\gamma}_{\text{node reconstruction}}, \gamma >1
\end{equation}
\begin{equation}
\label{eq:eq101}
\mathcal{L}_{edge\_recon}=\underbrace{\|A-\sigma(X \cdot X^T)\|_2^2}_{\text{edge reconstruction}}
\end{equation}
Wherein,  $\mathcal{L}_{node\_recon}$is the reconstruction loss for node features, and the reconstruction loss is measured using the cosine error. Furthermore, a scalable variant, the Scaled Cosine Error $\gamma$, is introduced to further refine the cosine error metric. $\mathcal{L}_{edge\_recon}$is the reconstruction loss for edges. Additionally, the term $KL(q_{\phi}(z|x) || p_{\theta}(z))$ embodies the Kullback-Leibler divergence, which serves to encourage the posterior distribution of codeword selections,$q_{\phi}(z|x)$, to closely align with the prior distribution $p_{\theta}(z)$, which is uniform.

\textbf{4.1.4 Other Methods for Selecting Vectors.  }A straightforward approach for selecting codebook vectors involves employing the argmax operation for nearest neighbor search. We conducted comparative experiments in Section 5.2 and found that using Gumbel Softmax has better performance. Additionally, we juxtaposed it with Gumbel-Argmax; however, both methods were found to inadequately capture a broader spectrum of graph structural variations, thereby underlining the advantage of the Gumbel-Softmax approach in retaining richer structural representations.

\subsection{LLM’s graph structure learning}

After Stage 1 , we acquire  the quantized embeddings and relaxed distributions representing both  node text and spatial information. Next, we utilize the same approach to obtain the two components of the LLM and harmonize the GNN’s modeling capacity in an aligned manner. As Figure~\ref{fig:main} Stage2

\textbf{4.2.1 Alignment of Graph Structure Modeling Ability.    }We input the nodes' linkage and textual information into the LLM and fine-tune it for the specific main task. Concurrently, we derive the quantized embeddings and relaxed distributions of the LLM utilizing the trained codebook according to Equations ~\ref{eq:eq5}\ref{eq:eq6}\ref{eq:eq7}. Subsequently, we perform alignment of the two components separately.

Formally, for each node, we obtain the embeddings of the last layer of the LLM, \( h_{llm} \), and the embedding extracted by the GNN, \( h_{gnn} \). Then, through Gumbel Softmax relaxation processing, we obtain the weights for different codewords, \( p_{llm} \) and \( p_{gnn} \), and the quantized embeddings, \( z_{llm} \) and \( z_{gnn} \) . 
\begin{equation}
\begin{aligned}
\label{eq:eq200}
z_{llm},p_{llm}=Gumbel\_softmax(h_{llm})\\
z_{gnn},p_{gnn}=Gumbel\_softmax(h_{gnn})
\end{aligned}
\end{equation}

The overall training loss of the LLM is
\begin{equation}
\label{eq:eq9}
\mathcal{L}_{Stage2}=\mathcal{L}_{CE}+\alpha_{mse} \mathcal{L}_{MSE}(z_{llm},z_{gnn})+ \beta_{kl} KL(p_{llm} || p_{gnn})
\end{equation}
Wherein, $\mathcal{L}_{CE}$ is the loss function for node classification of the LLM, $\mathcal{L}_{MSE}$  and $KL$ divergence are the loss functions for measuring the LLM and GNN in terms of quantized embeddings and relaxed distributions. $\alpha_{mse}$ and $\beta_{kl}$ are the corresponding hyperparameters.

\textbf{4.2.2 LLM inference.  } 
During the inference stage, our model diverges from existing approaches by dispensing with the necessity of employing GNNs for extracting structural information or integrating supplementary external data. It is capable of performing robust inference tasks using textual information and link information alone. We  successfully develop a codebook tailored to the graph data modality, thereby achieving alignment between the textual description of the graph and its topological structure in the LLM's processing and execution.
\subsection{Theoretical analysis}
Our objective centers on learning the optimal quantized embedding value, \(z_{llm}^*\), at the LLM end. Both \(z_{llm}\)  and \(z_{gnn}\) (the quantized embedding at GNN end) are representations derived from graph structural information, which is informed by prior knowledge \(y\). Specifically, we aim to learn \(z_{llm}^*\) by maximizing the conditional probability framework, refining these embeddings to encapsulate the structural essence conveyed by the graph under the guidance of \(y\).
\begin{equation}
e^*={\text{arg max}}\mathbb{E}_{p(z_{llm},z_{gnn})}[p(y,z_{gnn}|z_{llm})]
\end{equation}
\textbf{Theorem.  } Maximizing the posterior probability given prior information \(y\), represented as \(E_p(z_{llm}, z_{gnn})[p(y, z_{gnn}|z_{llm})]\), is equivalent to maximizing the mutual information \(I(z_{llm};z_{gnn})\) between the quantized embeddings \(z_{llm}\) from the LLM side and \(z_{gnn}\) from the GNN side.
 
\textbf{Proof.   } Notably, the training process unfolds in stages, wherein the quantized embedding $z_{gnn}$ from the GNN end is treated as static while optimizing the quantized embedding $z_{llm}$ at the LLM end.  This sequential approach permits us to derive the following insights:
\begin{equation}
\begin{aligned}
E_{p(z_{llm},z_{gnn})}[p(y,z_{gnn}|z_{llm})]\propto E_{p(z_{llm},z_{gnn})}\log\int_Z\frac{p(y,z_{gnn}|z_{llm})}{p(z_{gnn})}dz\\
=E_{p(z_{llm},z_{gnn})}\log\frac{\int_Z py,z_{llm}|z_{gnn})dz}{p(z_{llm})}\\
=E_{p(z_{llm},z_{gnn})}\log\frac{p(z_{llm}|z_{gnn})}{p(z_{llm})}=I(z_{llm},z_{gnn}).
\end{aligned}
\end{equation}

\section{EXPERIMENTS}
\textbf{Dataset Settings.  }We evaluate the proposed model using several widely used public benchmark datasets(Cora, Pubmed), and one OGB dataset(Arxiv, Arxiv-2023)~\cite{hu2020open}. And Arxiv-2023 is employed to avoid data leakage issues. Details can be found in Appendix ~\ref{sec:dataset}.

\textbf{Baselines.   } To evaluate the effectiveness of our proposed method, we compare it with several baselines from four categories. (1) Traditional GNN models including GCN~\cite{kipf2016semi}, GAT~\cite{velivckovic2017graph}, GraphSage~\cite{defferrard2016convolutional}, (2) Graph Transformers models such as Graphormers~\cite{yang2021graphformers} and NodeFormer~\cite{wu2022nodeformer}, (3) LM+GNN methods including GLEM~\cite{zhao2022learning} and GraphEdit~\cite{guo2024graphedit}, (4) LLM-based methods, which are further divided into two subcategories: raw text input (UniGraph~\cite{he2024unigraph}, InstructGLM~\cite{ye2023natural}, GIANT-XRT+GraphMAE2~\cite{hou2023graphmae2}) and embedding input (LLaGA~\cite{chen2024llaga}, GraphAdapter~\cite{huang2024can}, ENGINE~\cite{zhu2024efficient}). Details of these methods are in Appendix~\ref{sec:baseline}.

\textbf{Implementation Details.    }For fair comparison, we adopt  GraphSAGE as the GNN model and shallow embedding. We employ the Llama-2-7B model as the base model. In the embedding input category of LLM-based methods, we use TAPE. Additionally, we compare different embeddings in the Appendix~\ref{sec:embedding}. In terms of codebook training, we adjust the hyperparameters such as codebook dimension and size based on different datasets, with more details provided in the Appendix ~\ref{sec:hyperparameters}. The computing resource is NVIDIA RTX A800 80G.

\begin{table}[t]
\centering
\begin{tabular}{l|l|ccccc}

\toprule
\textbf{Type} & \textbf{Model} & \textbf{Arxiv} & \textbf{Pubmed} & \textbf{Cora} & \textbf{Arxiv2023} & \textbf{Average}\\
\midrule
\multirow{5}{*}{\shortstack[l]{Shallow embedding}}
& GCN & 0.7182 & 0.8031 & 0.8778 & 0.6760 & 0.7687 \\
& GraphSage & 0.7171 & 0.8881 & 0.8824 & 0.6906  & 0.7945\\
& GAT & 0.7366 & 0.8328 & 0.8595 & 0.6784 & 0.7768 \\
& RevGAT & 0.7402 & 0.8850 & 0.8911 & 0.6979  & 0.8035 \\
& DRGAT & 0.7416 & 0.8962 & 0.8977 & 0.7003 & 0.8089 \\
\cmidrule(lr){1-7}
\multirow{4}{*}{\shortstack[l]{Raw text(title)}}
& UniGraph & 0.7291 & 0.7433 & 0.8184 & ---  & --\\
& InstructGLM & 0.7297 & 0.9105 & 0.8977 & 0.7651 & 0.8258\\
& GraphMAE2 & 0.7201 & 0.6983 & 0.8011 & 0.7163 & 0.7340 \\
& LangTopo & 0.7365 & 0.9287 & 0.8998 & 0.7738 & 0.8347 \\
\cmidrule(lr){1-7}
\multirow{2}{*}{\shortstack[l]{GNN transformer}}
& Graphformer & 0.6725 & 0.7699 & 0.8044 & 0.6287 & 0.7188 \\
& Nodeformer & 0.6960 & 0.7958 & 0.8848 & 0.6744 & 0.7627 \\
\cmidrule(lr){1-7}
\multirow{2}{*}{\shortstack[l]{LLM enhance GNN}}
& GLEM & 0.7580 & 0.9459 & 0.8856 & 0.7858 & 0.8438\\
& GraphEdit & 0.7578 & 0.9409 & 0.9090 & 0.7966 & 0.8510\\
\cmidrule(lr){1-7}
\multirow{3}{*}{\shortstack[l]{Embedding by LLM}}
& LLaGA & 0.7666 & 0.9503 & 0.8922 & 0.8037 & 0.8532\\
& ENGINE & 0.7602 & 0.9477 & 0.9148 & 0.7976 & 0.8550\\
& LangTopo & \textbf{0.7681} & \textbf{0.9667} &  \textbf{0.9158} & \textbf{0.8126} & \textbf{0.8658}\\
\bottomrule
\end{tabular}
\caption{Main Results. Experiments were conducted comparing four categories of baseline models. The evaluation metric used was accuracy, with top-performing results emphasized \textbf{in bold}.}
\label{tab:main}
\end{table}

\subsection{Main Result}
We evaluate LangTopo against several baseline models on three public datasets, and the results are summarized in Table~\ref{tab:main}.
It is evident that LangTopo demonstrates remarkable performance across all datasets. Whether employing raw text as input or utilizing preprocessed embeddings, LangTopo consistently outperforms other models with LLM integration. Furthermore, we observe that using raw text directly yields inferior outcomes compared to using text embeddings. We attribute this discrepancy to the potential challenge LLMs face in handling lengthy texts, as raw texts tend to be extensive.

\subsection{Codebook Analysis}
In this subsection, we conduct a detailed analysis and comparison of the performance and effectiveness of the codebook.

\textbf{Different  Methods for selecting Vectors. 
 }Our experiments employed Gumbel softmax relaxation for the calculation  of quantized embeddings. Additionally, we compared the selection of quantized embeddings using Argmin cosine, euclidean distance, and Gumbel argmax. 
 
\begin{figure}[t]
    \centering
    \caption{
    The distribution of codebook embeddings with different strategies on the unit hypersphere.}
    \label{fig:codebookembedding}
    \includegraphics[width=1\textwidth]{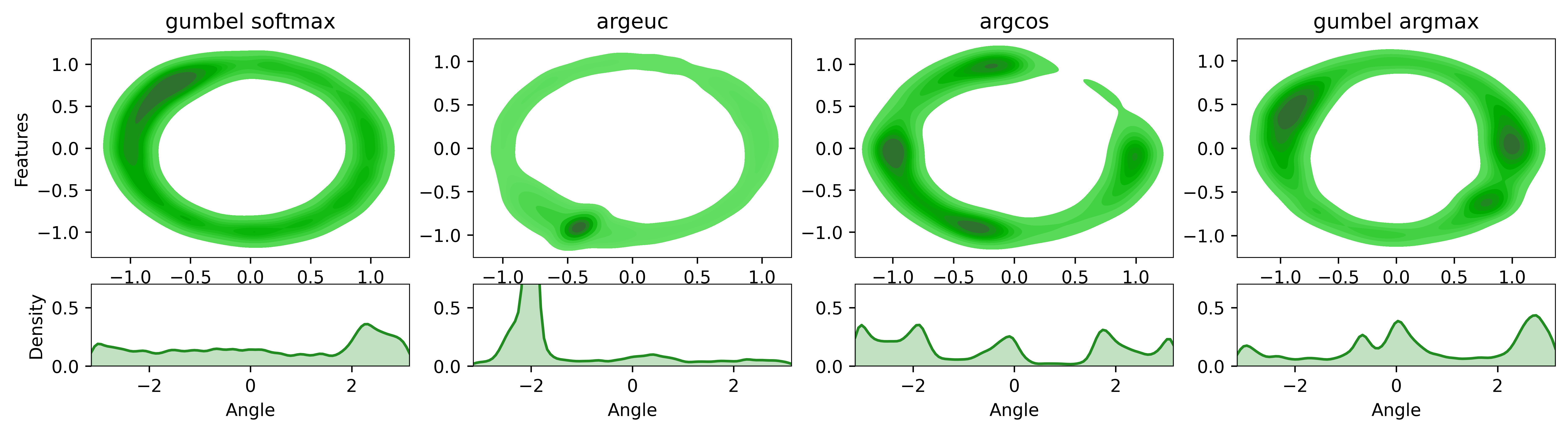}
\end{figure}

 Table~\ref{tab:table3} showcases the evaluation of codebooks constructed via differing strategies, revealing that Gumbel-softmax relaxation notably excels in creating a codebook with a high utilization rate. This characteristic is particularly advantageous for accurately modeling intricate graph datasets. In comparison, using argmax Euc yields the worst results, while the effects of cosine distance and Gumbel argmax are similar. In Figure~\ref{fig:codebookembedding}, we visualized the distribution of codebook embeddings with different strategies on the unit hypersphere, further demonstrating the effectiveness of the Gumbel softmax relaxation. Its embedding distribution is more uniform and exhibits better generalization performance.
 
\begin{table}[h]
  \caption{code perplexity and usage rate.Code perplexity represents the model’s average usage of the codebook, while the usage rate indicates the model’s utilization of the codebook. Generally, higher values are preferable.}
  \centering
  \label{tab:table3}
  \begin{tabular}{ccccc}
    \toprule
Methods&\textbf{Gumbel softmax}&\textbf{Gumbel argmax}&\textbf{argmax Cos }&\textbf{argmax Euc}\\
    \midrule
    \#perplexity & 2263 & 1230 & 1168& 778\\
    \#usage & 0.7324 & 0.493& 0.501&0.323 \\
  \bottomrule
\end{tabular}
\end{table}
\subsection{Ablation Study}

\begin{figure}[h]
    \centering
    \includegraphics[width=1\linewidth]{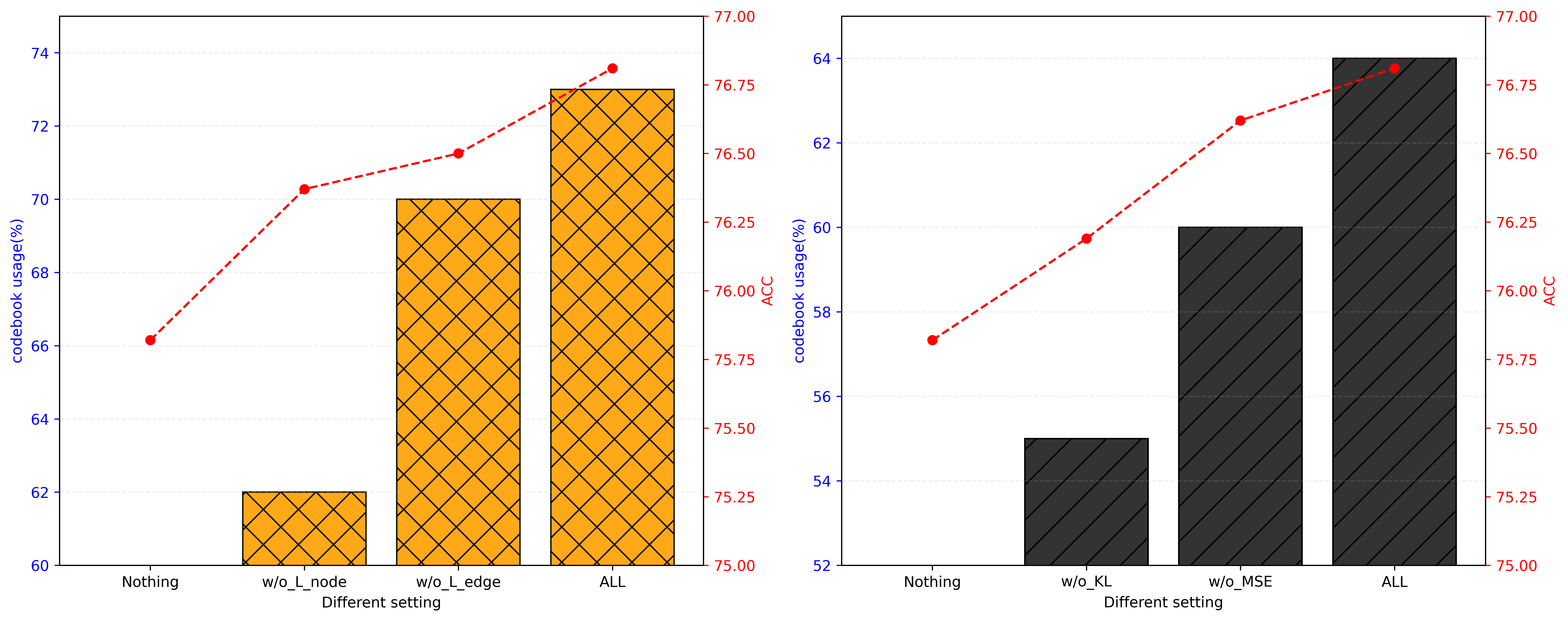}
    \caption{Our investigation into diverse loss functions within the LangTopy architecture has substantiated the importance and efficacy of individual loss functions. The left figure examines the efficacy of node reconstruction and edge reconstruction loss functions, while the right figure delves into the importance of relaxed distributions and quantized embeddings in the learning process of LLMs.}
    \label{fig:loss_abtion}
\end{figure}

\textbf{Loss Ablation.   }To delve deeper into understanding the influence of different loss functions on codebook construction and enhance the graph structural modeling capability of LLMs, we conducted an ablation study on the arxiv dataset, as shown in Figure~\ref{fig:loss_abtion} .

As evident from the table, each loss function in the LangTopo structure plays a crucial role, and we observe a positive correlation between the improved codebook utilization by LLMs and the accuracy in LLM node classification, further validating the correctness of the design direction of the LangTopo architecture.

\textbf{Effect of different Hop.   }
We conducted an ablation study to investigate the contributions of our varying hop ranges to LangTopo's performance. To this end, we examined the impact of different hop ranges – 0-hop, 1-hop, and 2-hop – on LangTopo's efficacy across three datasets, as well as the contribution of these different hop ranges to LLMs' graph task-solving abilities without the LangTopo framework. The results, visible in Table~\ref{tab:hop}, lead us to the following conclusions:

(i) Compared to scenarios without LangTopo, the use of LangTopo yields superior performance in both 1-hop and 2-hop settings. However, in the 0-hop configuration, LangTopo does not exhibit any difference in effect, which aligns with expectations. Because LangTopo cannot model graph structure effectively without structural information, as it is inherently designed to leverage such details.

(ii) LangTopo achieves its best performance in the 2-hop setting, whereas the without LangTopo only attains optimal results in the 1-hop scenario. These experiments further substantiate that LangTopo enhances the LLM's capability to model graph structures.
\begin{table}[h]
\centering
\begin{tabular}{l|l|ccc}
\toprule
Methods & hop & ARXIV & Pubmed & Cora \\
\midrule
w/o LangTopo & 0-hop   & 73.88 & 94.09 & 85.64\\
      & 1-hop & 76.01 & 94.60 & 89.31\\
      & 2-hop & 75.78 & 94.01 & 89.12\\
\midrule
w LangTopo & 0-hop   & 73.85 & 94.17 & 85.49\\
       & 1-hop & 76.64 & 96.53 & 91.45\\
       & 2-hop & 76.81 & 96.67 & 91.58\\
\bottomrule
\end{tabular}
\caption{Effect comparison under different hop counts between with LangTopo and without LangTopo.}
\label{tab:hop}
\label{tab:embedding-node-classification-results}
\end{table}

\textbf{Effect of different LLM.   }In our main experiment, Llama-2-7B serves as the foundational model. We extend our study beyond Llama-2-7B, considering alternative models such as Vicuna-7B~\cite{chiang2023vicuna}, and OPT-2.7B~\cite{zhang2022opt}. The results after substituting these models are presented in Table~\ref{tab:llm}
\begin{table}[h]
\centering
\caption{Performance comparison using different LLM.}
\label{tab:llm}
\begin{tabular}{lccc}
\toprule
LLM & Arxiv & Pubmed & Cora \\
\midrule
OPT-2.7B & 0.7677 & 0.9629 & 0.9184\\
vicuna-7B & 0.7679 & 0.9654 & 0.9177 \\
Llama-2-7B & 0.7681 & 0.9667 & 0.9158 \\
\bottomrule
\end{tabular}
\end{table}

\textbf{Others.   }We assess the impact of different embeddings in Appendix~\ref{sec:embedding}, examine the influence of selecting different GNNs on model performance in Appendix~\ref{sec:GNN}, provide details on prompt design in Appendix~\ref{sec:prompt}, and investigate the differing outcomes of rich versus minimal text in Appendix~\ref{sec:text}.

%\textbf{Rich text and Raw text.   }
\section{CONCLUSION}
We present an innovative graph structural learning framework, LangTopo, tailored for LLMs. This framework harnesses an approach to quantify the graph modeling capacities of GNNs and LLMs, thereby achieving a substantial enhancement in LLMs' comprehension of graph topologies. Moreover, LangTopo transcends the limitations of prevailing paradigms that necessitate external models to apprehend graph structures during the inference phase. Standing as a pioneering work, LangTopo is poised to furnish a robust benchmark for guiding future advancements within this domain.

\textbf{Limitation.   }A limitation in our experimental setup is the unexplored scenario of jointly training with multiple datasets for graph modality codebooks. Presently, our evaluations are confined to individual datasets. In future, we intend to investigate the generalizability and scalability of LangTopo.

\newpage
\bibliographystyle{ACM-Reference-Format}
\bibliography{sample-base}

\appendix % 标记开始进入附录部分

\newpage

\section{Baseline}
\label{sec:baseline}
Traditional GNNs: This work employs five simple yet widely-used GNN models, namely GCN~\cite{kipf2016semi}, GraphSAGE~\cite{van2017neural}, GAT~\cite{velivckovic2017graph}, DRGAT~\cite{li2021training}, and RevGAT~\cite{luan2022revisiting}.

GraphFormers~\cite{yang2021graphformers}: These are graph transformers with nested GNN layers originally designed for link prediction tasks.

NodeFormer~\cite{wu2022nodeformer}: It is an efficient large-graph transformer tool that devises a kernelized Gumbel-Softmax operator.

GLEM~\cite{sun2023large}: This is an effective framework, integrating large language models (LLMs) with GNNs via a variational EM framework during training.

GraphEdit~\cite{guo2024graphedit}: It enhances the reliability of graph structure learning by effectively denoising noisy inputs through the leveraging of LLMs.

UniGraph~\cite{he2024unigraph}: It adopts a masking strategy to facilitate joint training of LLMs and GNNs under cascaded structures, enabling generalization to other graphs and datasets.

InstructGLM~\cite{ye2023natural}: It utilizes LLM-based instruction fine-tuning to describe the geometric structure of graphs and node features, tackling graph tasks thereby.

LLaGA~\cite{chen2024llaga}: It reformats link information between nodes into sequential data, feeding LLM-compatible sequential inputs while preserving node structure, achieved through instruction fine-tuning.

ENGINE~\cite{zhu2024efficient}: By devising tunable side structures, it combines LLMs with GNNs and employs various strategies to expedite model training and inference.

\section{Effect of different Embedding}
\label{sec:embedding}
To evaluate the impact of different embeddings on model performance, we selected  OGB embeddings and GIANT embeddings for comparison with various GNN models, and the results are as shown in the table. It can be observed that LangTopo consistently outperforms GNN models across different embedding scenarios.
\begin{table}[h]
\centering
\caption{Performance comparison using different embeddings(arxiv).}
\label{tab:embedding}
\begin{tabular}{lcccccc}
\toprule
LM & GCN & SAGE & GAT & RevGAT & DRGAT & LangTopo \\
\midrule
OGB& 0.7174 & 0.7119 & 0.7366 & 0.7402 & 0.7416 & \textbf{0.7451}\\
GIANT & 0.7329 & 0.7435 & 0.7416 & 0.7590 & 0.7611 & \textbf{0.7624}\\
\bottomrule
\end{tabular}
\end{table}

\section{rich and scarce text}
\label{sec:text}
We have delved into the capacity of LLMs to utilize textual information, categorizing it into two primary types: 1) Rich Text, comprising both summaries and titles, and 2) Scarce Text, which is limited to titles only. The outcomes, as depicted in the table~\ref{tab:text}, indicate that LLMs experience a notable enhancement when furnished with rich text.
\begin{table}[h]
\centering
\caption{Performance comparison between 
rich text(abstract title) and scarce text(title). raw text}
\label{tab:text}
\begin{tabular}{lccc}
\toprule
LLM & Arxiv & Pubmed & Cora \\
\midrule
rich & 0.7598 & 0.9395 & 0.9085\\
title & 0.7365 & 0.9287 & 0.8998 \\
\bottomrule
\end{tabular}
\end{table}

\section{Dataset}
\label{sec:dataset}
We have compiled statistical measures for each dataset, as illustrated in the table~\ref{tab:table1} provided. Furthermore, we employed the arxiv2023 dataset, which compiles scholarly articles from the arXiv website for the year 2023. This strategic choice was made to circumvent potential data leakage issues arising from the use of corpus included in the dataset during the pre-training phase of the LLMs.
\begin{table}[h]
  \caption{Statistics of the evaluation datasets.}
  \centering
  \label{tab:table1}
  \begin{tabular}{ccccc}
    \toprule
Dataset&\textbf{Arxiv}&\textbf{Pubmed}&\textbf{Cora}&\textbf{arxiv23}\\
    \midrule
    \#Nodes & 169343 &19717 &2708&46198\\
    \#edges & 1166243 &44338 &5429&78548\\
    \#Classes & 40 &3 &7&40\\
    \#Split ratio(\%) & 54/18/28& 60/20/20&60/20/20&60/20/20\\
  \bottomrule
\end{tabular}
\end{table}

\section{Hyperparameters}
\label{sec:hyperparameters}

Table~\ref{sec:hyperparameters} furnishes the hyperparameters for LangTopo across various datasets. For more detailed configurations, please refer to our source code.
\begin{table}[h]
\caption{Hyperparameters for the Arxiv, Pubmed, and Citeseer.}
\label{tab:hyper}
\centering
\begin{tabular}{lccc}
\toprule
Hyperparameters & Arixv & Pubmed & Citeseer \\
\midrule
\textbf{GNN}\\
layers & 3 & 2 & 2 \\
hidden dim & 512 & 256 & 256 \\
learning rate & 0.001 & 0.005 & 0.005 \\
dropout & 0.5 & 0.5 & 0.5 \\
epoch & 1000 & 1000 & 1000 \\
early stop & 50 & 50 & 50 \\
codebook size& 4096 & 2048 & 2048 \\
$\alpha_{node}$ & 10 & 10 & 10 \\
$\beta_{KL}$ & 1e-2 & 1e-2 & 1e-2 \\
\textbf{LLM}\\
learning rate & 5e-5 & 5e-5 & 5e-5 \\
warmup & 0.05 & 0.05 & 0.05 \\
gradient accumulation steps & 8 & 8 & 8 \\
batch size & 4 & 4 & 4 \\
$\alpha_{mse}$ & 0.5 & 0.1 & 0.1 \\
$\beta_{kl}$& 1e-4 & 1e-4 & 1e-4 \\
\bottomrule
\end{tabular}
\end{table}

\section{Effect of different GNN}
\label{sec:GNN}
We conducted an exploration of LangTopo by substituting it with different GNN models.
\begin{table}[h]
\centering
\caption{Performance comparison using different GNN.}
\label{tab:llm}
\begin{tabular}{lccc}
\toprule
LLM & Arxiv & Pubmed & Cora \\
\midrule
GCN & 0.7640 & 0.9566 & 0.9012\\
GraphSage & 0.7681 & 0.9667 & 0.9158\\
GAT & 0.7704 & 0.9633 & 0.9123 \\
\bottomrule
\end{tabular}
\end{table}

\section{Gumbel softmax}
We will show that using gumbel softmax does not change the selection probability distribution.
\label{sec:appendixA}
\noindent Consider a \(K\)-dimensional output vector where each component is denoted \(x_k\). The probability of obtaining each dimension via the softmax function is:

\begin{equation}
pi_k = \frac{e^{x_k}}{\sum_{k'=1}^K e^{x_{k'}}}
\end{equation}
\noindent If we add independent standard Gumbel noise (with scale parameter 1 and location parameter 0) to each \(x_k\), and select the dimension corresponding to the maximum value as the output, it is claimed that the resulting probability remains \(\pi_k\). We proceed to prove this assertion.
The probability density function (PDF) of a Gumbel distribution (with scale 1 and location \(\mu\)) is:
\[
f(z;\mu) = e^{-(z-\mu) - e^{-(z-\mu)}},
\]
and its cumulative distribution function (CDF) is:
\[
F(z;\mu) = e^{-e^{-(z-\mu)}}.
\]
Assuming the \(k\)-th Gumbel distribution corresponds to \(x_k\), yielding \(z_k = x_k + G_k\), where \(G_k\) follows a Gumbel distribution with location parameter \(x_k\), we aim to show that the probability of obtaining \(z_k\) is indeed \(\pi_k\), i.e., 
\[
P(z_k \geq z_{k'}, \, \forall k' \neq k | \{x_{k'}\}_{k'=1}^K) = \pi_k.
\]
The conditional cumulative probability for \(z_k\) is then:
\[
P(z_k \geq z_{k'}, \, \forall k' \neq k | z_k, \{x_{k'}\}_{k'=1}^K) = \prod_{k' \neq k} e^{-e^{-(z_k - x_{k'})}}.
\]

Integrating over \(z_k\) to obtain the marginal probability yields:
\begin{equation}
\begin{aligned}
P(z_k \geq z_{k'}, \, \forall k' \neq k | \{x_{k'}\}_{k'=1}^K) \\
&= \int P(z_k \geq z_{k'}, \, \forall k' \neq k | z_k, \{x_{k'}\}_{k'=1}^K) \cdot f(z_k; x_k) \, dz_k \\
&= \int \prod_{k' \neq k} e^{-e^{-(z_k - x_{k'})} } \cdot e^{-(z_k - x_k) - e^{-(z_k - x_k)}} \, dz_k \\
&= \int e^{-\sum_{k' \neq k} e^{-(z_k - x_{k'})} - (z_k - x_k) - e^{-(z_k - x_k)}} \, dz_k \\
&= \int e^{-\sum_{k' \neq k} e^{-(z_k - x_{k'})} - (z_k - x_k) - e^{-(z_k - x_k)}} \, dz_k \\
&= \int e^{-\sum_{k'=1}^K e^{-(z_k - x_{k'})} - z_k + x_k} \, dz_k \\
&= \int e^{-e^{-z_k + \ln\left(\sum_{k'=1}^K e^{x_{k'}}\right)} - z_k + x_k} \, dz_k \\
&= e^{-\ln\left(\sum_{k'=1}^K e^{x_{k'}}\right) + x_k} \int e^{-e^{-(z_k - \ln\left(\sum_{k'=1}^K e^{x_{k'}}\right))} - (z_k - \ln\left(\sum_{k'=1}^K e^{x_{k'}}\right))} \, dz_k \\
&= \frac{e^{x_k}}{\sum_{k'=1}^K e^{x_{k'}}} \int e^{-e^{-(z_k - \ln\left(\sum_{k'=1}^K e^{x_{k'}}\right))} - (z_k - \ln\left(\sum_{k'=1}^K e^{x_{k'}}\right))} \, dz_k \\
&= \frac{e^{x_k}}{\sum_{k'=1}^K e^{x_{k'}}} \int {-e^{-(z_k - \ln\left(\sum_{k'=1}^K e^{x_{k'}}\right))} - e^{(z_k - \ln\left(\sum_{k'=1}^K e^{x_{k'}}\right))}} \, dz_k \\
\end{aligned}
\end{equation}
Inside the integral, we consider a Gumbel distribution characterized by \(\mu = \ln\left(\sum_{k'=1}^K e^{x_{k'}}\right)\), from which it is known that the integral evaluates to 1. Consequently, this leads us to the conclusion that
\[
P(z_k \geq z_{k'}, \, \forall k' \neq k \mid \{x_{k'}\}_{k'=1}^K) = \frac{e^{x_k}}{\sum_{k'=1}^K e^{x_{k'}}},
\]
which is consistent with the output of the softmax function.

\newpage
\section{Prompt in different dataset}
\label{sec:prompt}

\begin{table*}[h]
	\centering
	\begin{tabularx}{\textwidth}{lXXX}
		\toprule
		&  Prompt \\
		\midrule
		Cora 
		& "<User>: Given a node \{\textcolor{blue}{text attribute}\}. the node connect \{\textcolor{blue}{link information}\}. Categorize the nodes into groups 'theory','reinforcement learning', 'neural networks', 'probabilistic methods', 'case based', and 'rule learning', according to the given information. the node \{\textcolor{blue}{text attribute}\} is classified as <Assistant>: \{\textcolor{blue}{label}\}" \\
  	Pubmed 
		& "<User>: Given a node \{\textcolor{blue}{text attribute}\}. the node connect \{\textcolor{blue}{link information}\}. Categorize the nodes into groups 'Diabetes Mellitus, Experimental','Diabetes Mellitus, Type 2', and 'Diabetes Mellitus, Type 1' according to the given information. the node  \{\textcolor{blue}{text attribute}\} is classified as <Assistant>: \{\textcolor{blue}{label}\}" \\
  	Arxiv 
		& "<User>: Given a node \{\textcolor{blue}{text attribute}\}, the node connect \{\textcolor{blue}{link information}\}. We need to classify the node into 40 classes: 'Numerical Analysis','Multimedia','Logic in Computer Science','Computers and Society','Cryptography and Security','Distributed, Parallel, and Cluster Computing','Human-Computer Interaction','Computational Engineering, Finance, and Science','Networking and Internet Architecture','Computational Complexity','Artificial Intelligence','Multiagent Systems','General Literature','Neural and Evolutionary Computing','Symbolic Computation','Hardware Architecture','Computer Vision and Pattern Recognition','Graphics','Emerging Technologies','Systems and Control','Computational Geometry','Other Computer Science','Programming Languages','Software Engineering','Machine Learning','Sound','Social and Information Networks','Robotics','Information Theory','Performance','Computation and Language','Information Retrieval','Mathematical Software','Formal Languages and Automata Theory','Data Structures and Algorithms','Operating Systems','Computer Science and Game Theory','Databases','Digital Libraries','Discrete Mathematics'. the node \{\textcolor{blue}{text attribute}\} is classified as <Assistant>: \{\textcolor{blue}{label}\}" \\
  	Arxiv2023 
		& "<User>: Given a node \{\textcolor{blue}{text attribute}\}, the node connect \{\textcolor{blue}{link information}\}. We need to classify the node into 40 classes: 'Numerical Analysis','Multimedia','Logic in Computer Science','Computers and Society','Cryptography and Security','Distributed, Parallel, and Cluster Computing','Human-Computer Interaction','Computational Engineering, Finance, and Science','Networking and Internet Architecture','Computational Complexity','Artificial Intelligence','Multiagent Systems','General Literature','Neural and Evolutionary Computing','Symbolic Computation','Hardware Architecture','Computer Vision and Pattern Recognition','Graphics','Emerging Technologies','Systems and Control','Computational Geometry','Other Computer Science','Programming Languages','Software Engineering','Machine Learning','Sound','Social and Information Networks','Robotics','Information Theory','Performance','Computation and Language','Information Retrieval','Mathematical Software','Formal Languages and Automata Theory','Data Structures and Algorithms','Operating Systems','Computer Science and Game Theory','Databases','Digital Libraries','Discrete Mathematics'. the node \{\textcolor{blue}{text attribute}\} is classified as <Assistant>: \{\textcolor{blue}{label}\}" \\
		\bottomrule
	\end{tabularx}%
	\label{tab:addlabel}%
	\caption{A table with different prompt}
\end{table*}%

\newpage
\section*{NeurIPS Paper Checklist}

\begin{enumerate}

\item {\bf Claims}
    \item[] Question: Do the main claims made in the abstract and introduction accurately reflect the paper's contributions and scope?
    \item[] Answer: \answerYes{} % Replace by \answerYes{}, \answerNo{}, or \answerNA{}.
    \item[] Justification: The abstract and introduction have accurately expressed the topic of the paper
    \item[] Guidelines:
    \begin{itemize}
        \item The answer NA means that the abstract and introduction do not include the claims made in the paper.
        \item The abstract and/or introduction should clearly state the claims made, including the contributions made in the paper and important assumptions and limitations. A No or NA answer to this question will not be perceived well by the reviewers. 
        \item The claims made should match theoretical and experimental results, and reflect how much the results can be expected to generalize to other settings. 
        \item It is fine to include aspirational goals as motivation as long as it is clear that these goals are not attained by the paper. 
    \end{itemize}

\item {\bf Limitations}
    \item[] Question: Does the paper discuss the limitations of the work performed by the authors?
    \item[] Answer: \answerYes{} % Replace by \answerYes{}, \answerNo{}, or \answerNA{}.
    \item[] Justification: The relevant limitations of the paper have been discussed, and the relevant exploration has been insufficient.
    \item[] Guidelines:
    \begin{itemize}
        \item The answer NA means that the paper has no limitation while the answer No means that the paper has limitations, but those are not discussed in the paper. 
        \item The authors are encouraged to create a separate "Limitations" section in their paper.
        \item The paper should point out any strong assumptions and how robust the results are to violations of these assumptions (e.g., independence assumptions, noiseless settings, model well-specification, asymptotic approximations only holding locally). The authors should reflect on how these assumptions might be violated in practice and what the implications would be.
        \item The authors should reflect on the scope of the claims made, e.g., if the approach was only tested on a few datasets or with a few runs. In general, empirical results often depend on implicit assumptions, which should be articulated.
        \item The authors should reflect on the factors that influence the performance of the approach. For example, a facial recognition algorithm may perform poorly when image resolution is low or images are taken in low lighting. Or a speech-to-text system might not be used reliably to provide closed captions for online lectures because it fails to handle technical jargon.
        \item The authors should discuss the computational efficiency of the proposed algorithms and how they scale with dataset size.
        \item If applicable, the authors should discuss possible limitations of their approach to address problems of privacy and fairness.
        \item While the authors might fear that complete honesty about limitations might be used by reviewers as grounds for rejection, a worse outcome might be that reviewers discover limitations that aren't acknowledged in the paper. The authors should use their best judgment and recognize that individual actions in favor of transparency play an important role in developing norms that preserve the integrity of the community. Reviewers will be specifically instructed to not penalize honesty concerning limitations.
    \end{itemize}

\item {\bf Theory Assumptions and Proofs}
    \item[] Question: For each theoretical result, does the paper provide the full set of assumptions and a complete (and correct) proof?
    \item[] Answer: \answerYes{}
    \item[] Justification: Hypotheses and proofs are provided in the body and appendix
    \item[] Guidelines:
    \begin{itemize}
        \item The answer NA means that the paper does not include theoretical results. 
        \item All the theorems, formulas, and proofs in the paper should be numbered and cross-referenced.
        \item All assumptions should be clearly stated or referenced in the statement of any theorems.
        \item The proofs can either appear in the main paper or the supplemental material, but if they appear in the supplemental material, the authors are encouraged to provide a short proof sketch to provide intuition. 
        \item Inversely, any informal proof provided in the core of the paper should be complemented by formal proofs provided in appendix or supplemental material.
        \item Theorems and Lemmas that the proof relies upon should be properly referenced. 
    \end{itemize}

    \item {\bf Experimental Result Reproducibility}
    \item[] Question: Does the paper fully disclose all the information needed to reproduce the main experimental results of the paper to the extent that it affects the main claims and/or conclusions of the paper (regardless of whether the code and data are provided or not)?
    \item[] Answer: \answerYes{} % Replace by \answerYes{}, \answerNo{}, or \answerNA{}.
    \item[] Justification: We provide the data and code
    \item[] Guidelines:
    \begin{itemize}
        \item The answer NA means that the paper does not include experiments.
        \item If the paper includes experiments, a No answer to this question will not be perceived well by the reviewers: Making the paper reproducible is important, regardless of whether the code and data are provided or not.
        \item If the contribution is a dataset and/or model, the authors should describe the steps taken to make their results reproducible or verifiable. 
        \item Depending on the contribution, reproducibility can be accomplished in various ways. For example, if the contribution is a novel architecture, describing the architecture fully might suffice, or if the contribution is a specific model and empirical evaluation, it may be necessary to either make it possible for others to replicate the model with the same dataset, or provide access to the model. In general. releasing code and data is often one good way to accomplish this, but reproducibility can also be provided via detailed instructions for how to replicate the results, access to a hosted model (e.g., in the case of a large language model), releasing of a model checkpoint, or other means that are appropriate to the research performed.
        \item While NeurIPS does not require releasing code, the conference does require all submissions to provide some reasonable avenue for reproducibility, which may depend on the nature of the contribution. For example
        \begin{enumerate}
            \item If the contribution is primarily a new algorithm, the paper should make it clear how to reproduce that algorithm.
            \item If the contribution is primarily a new model architecture, the paper should describe the architecture clearly and fully.
            \item If the contribution is a new model (e.g., a large language model), then there should either be a way to access this model for reproducing the results or a way to reproduce the model (e.g., with an open-source dataset or instructions for how to construct the dataset).
            \item We recognize that reproducibility may be tricky in some cases, in which case authors are welcome to describe the particular way they provide for reproducibility. In the case of closed-source models, it may be that access to the model is limited in some way (e.g., to registered users), but it should be possible for other researchers to have some path to reproducing or verifying the results.
        \end{enumerate}
    \end{itemize}

\item {\bf Open access to data and code}
    \item[] Question: Does the paper provide open access to the data and code, with sufficient instructions to faithfully reproduce the main experimental results, as described in supplemental material?
    \item[] Answer: \answerYes{} % Replace by \answerYes{}, \answerNo{}, or \answerNA{}.
    \item[] Justification: We provide the corresponding data and code in zip file
    \item[] Guidelines:
    \begin{itemize}
        \item The answer NA means that paper does not include experiments requiring code.
        \item Please see the NeurIPS code and data submission guidelines (\url{https://nips.cc/public/guides/CodeSubmissionPolicy}) for more details.
        \item While we encourage the release of code and data, we understand that this might not be possible, so “No” is an acceptable answer. Papers cannot be rejected simply for not including code, unless this is central to the contribution (e.g., for a new open-source benchmark).
        \item The instructions should contain the exact command and environment needed to run to reproduce the results. See the NeurIPS code and data submission guidelines (\url{https://nips.cc/public/guides/CodeSubmissionPolicy}) for more details.
        \item The authors should provide instructions on data access and preparation, including how to access the raw data, preprocessed data, intermediate data, and generated data, etc.
        \item The authors should provide scripts to reproduce all experimental results for the new proposed method and baselines. If only a subset of experiments are reproducible, they should state which ones are omitted from the script and why.
        \item At submission time, to preserve anonymity, the authors should release anonymized versions (if applicable).
        \item Providing as much information as possible in supplemental material (appended to the paper) is recommended, but including URLs to data and code is permitted.
    \end{itemize}

\item {\bf Experimental Setting/Details}
    \item[] Question: Does the paper specify all the training and test details (e.g., data splits, hyperparameters, how they were chosen, type of optimizer, etc.) necessary to understand the results?
    \item[] Answer: \answerYes{} % Replace by \answerYes{}, \answerNo{}, or \answerNA{}.
    \item[] Justification: We provide the corresponding data and code in zip file
    \item[] Guidelines:
    \begin{itemize}
        \item The answer NA means that the paper does not include experiments.
        \item The experimental setting should be presented in the core of the paper to a level of detail that is necessary to appreciate the results and make sense of them.
        \item The full details can be provided either with the code, in appendix, or as supplemental material.
    \end{itemize}

\item {\bf Experiment Statistical Significance}
    \item[] Question: Does the paper report error bars suitably and correctly defined or other appropriate information about the statistical significance of the experiments?
    \item[] Answer: \answerNo{} % Replace by \answerYes{}, \answerNo{}, or \answerNA{}.
    \item[] Justification: For LLM, the effect of repeated experiments is stable
    \item[] Guidelines:
    \begin{itemize}
        \item The answer NA means that the paper does not include experiments.
        \item The authors should answer "Yes" if the results are accompanied by error bars, confidence intervals, or statistical significance tests, at least for the experiments that support the main claims of the paper.
        \item The factors of variability that the error bars are capturing should be clearly stated (for example, train/test split, initialization, random drawing of some parameter, or overall run with given experimental conditions).
        \item The method for calculating the error bars should be explained (closed form formula, call to a library function, bootstrap, etc.)
        \item The assumptions made should be given (e.g., Normally distributed errors).
        \item It should be clear whether the error bar is the standard deviation or the standard error of the mean.
        \item It is OK to report 1-sigma error bars, but one should state it. The authors should preferably report a 2-sigma error bar than state that they have a 96\% CI, if the hypothesis of Normality of errors is not verified.
        \item For asymmetric distributions, the authors should be careful not to show in tables or figures symmetric error bars that would yield results that are out of range (e.g. negative error rates).
        \item If error bars are reported in tables or plots, The authors should explain in the text how they were calculated and reference the corresponding figures or tables in the text.
    \end{itemize}

\item {\bf Experiments Compute Resources}
    \item[] Question: For each experiment, does the paper provide sufficient information on the computer resources (type of compute workers, memory, time of execution) needed to reproduce the experiments?
    \item[] Answer: \answerYes{} % Replace by \answerYes{}, \answerNo{}, or \answerNA{}.
    \item[] Justification: We provided the appropriate information
    \item[] Guidelines:
    \begin{itemize}
        \item The answer NA means that the paper does not include experiments.
        \item The paper should indicate the type of compute workers CPU or GPU, internal cluster, or cloud provider, including relevant memory and storage.
        \item The paper should provide the amount of compute required for each of the individual experimental runs as well as estimate the total compute. 
        \item The paper should disclose whether the full research project required more compute than the experiments reported in the paper (e.g., preliminary or failed experiments that didn't make it into the paper). 
    \end{itemize}
    
\item {\bf Code Of Ethics}
    \item[] Question: Does the research conducted in the paper conform, in every respect, with the NeurIPS Code of Ethics \url{https://neurips.cc/public/EthicsGuidelines}?
    \item[] Answer: \answerYes{} % Replace by \answerYes{}, \answerNo{}, or \answerNA{}.
    \item[] Justification: Our experiment is ethical
    \item[] Guidelines:
    \begin{itemize}
        \item The answer NA means that the authors have not reviewed the NeurIPS Code of Ethics.
        \item If the authors answer No, they should explain the special circumstances that require a deviation from the Code of Ethics.
        \item The authors should make sure to preserve anonymity (e.g., if there is a special consideration due to laws or regulations in their jurisdiction).
    \end{itemize}

\item {\bf Broader Impacts}
    \item[] Question: Does the paper discuss both potential positive societal impacts and negative societal impacts of the work performed?
    \item[] Answer: \answerNA{} % Replace by \answerYes{}, \answerNo{}, or \answerNA{}.
    \item[] Justification: Our experiment didn't make a difference
    \item[] Guidelines:
    \begin{itemize}
        \item The answer NA means that there is no societal impact of the work performed.
        \item If the authors answer NA or No, they should explain why their work has no societal impact or why the paper does not address societal impact.
        \item Examples of negative societal impacts include potential malicious or unintended uses (e.g., disinformation, generating fake profiles, surveillance), fairness considerations (e.g., deployment of technologies that could make decisions that unfairly impact specific groups), privacy considerations, and security considerations.
        \item The conference expects that many papers will be foundational research and not tied to particular applications, let alone deployments. However, if there is a direct path to any negative applications, the authors should point it out. For example, it is legitimate to point out that an improvement in the quality of generative models could be used to generate deepfakes for disinformation. On the other hand, it is not needed to point out that a generic algorithm for optimizing neural networks could enable people to train models that generate Deepfakes faster.
        \item The authors should consider possible harms that could arise when the technology is being used as intended and functioning correctly, harms that could arise when the technology is being used as intended but gives incorrect results, and harms following from (intentional or unintentional) misuse of the technology.
        \item If there are negative societal impacts, the authors could also discuss possible mitigation strategies (e.g., gated release of models, providing defenses in addition to attacks, mechanisms for monitoring misuse, mechanisms to monitor how a system learns from feedback over time, improving the efficiency and accessibility of ML).
    \end{itemize}
    
\item {\bf Safeguards}
    \item[] Question: Does the paper describe safeguards that have been put in place for responsible release of data or models that have a high risk for misuse (e.g., pretrained language models, image generators, or scraped datasets)?
    \item[] Answer: \answerNA{} % Replace by \answerYes{}, \answerNo{}, or \answerNA{}.
    \item[] Justification: the paper poses no such risks.
    \item[] Guidelines:
    \begin{itemize}
        \item The answer NA means that the paper poses no such risks.
        \item Released models that have a high risk for misuse or dual-use should be released with necessary safeguards to allow for controlled use of the model, for example by requiring that users adhere to usage guidelines or restrictions to access the model or implementing safety filters. 
        \item Datasets that have been scraped from the Internet could pose safety risks. The authors should describe how they avoided releasing unsafe images.
        \item We recognize that providing effective safeguards is challenging, and many papers do not require this, but we encourage authors to take this into account and make a best faith effort.
    \end{itemize}

\item {\bf Licenses for existing assets}
    \item[] Question: Are the creators or original owners of assets (e.g., code, data, models), used in the paper, properly credited and are the license and terms of use explicitly mentioned and properly respected?
    \item[] Answer: \answerYes{} % Replace by \answerYes{}, \answerNo{}, or \answerNA{}.
    \item[] Justification: We followed the relevant protocols
    \item[] Guidelines:
    \begin{itemize}
        \item The answer NA means that the paper does not use existing assets.
        \item The authors should cite the original paper that produced the code package or dataset.
        \item The authors should state which version of the asset is used and, if possible, include a URL.
        \item The name of the license (e.g., CC-BY 4.0) should be included for each asset.
        \item For scraped data from a particular source (e.g., website), the copyright and terms of service of that source should be provided.
        \item If assets are released, the license, copyright information, and terms of use in the package should be provided. For popular datasets, \url{paperswithcode.com/datasets} has curated licenses for some datasets. Their licensing guide can help determine the license of a dataset.
        \item For existing datasets that are re-packaged, both the original license and the license of the derived asset (if it has changed) should be provided.
        \item If this information is not available online, the authors are encouraged to reach out to the asset's creators.
    \end{itemize}

\item {\bf New Assets}
    \item[] Question: Are new assets introduced in the paper well documented and is the documentation provided alongside the assets?
    \item[] Answer: \answerYes{}, % Replace by \answerYes{}, \answerNo{}, or \answerNA{}.
    \item[] Justification: We made the relevant Settings
    \item[] Guidelines:
    \begin{itemize}
        \item The answer NA means that the paper does not release new assets.
        \item Researchers should communicate the details of the dataset/code/model as part of their submissions via structured templates. This includes details about training, license, limitations, etc. 
        \item The paper should discuss whether and how consent was obtained from people whose asset is used.
        \item At submission time, remember to anonymize your assets (if applicable). You can either create an anonymized URL or include an anonymized zip file.
    \end{itemize}

\item {\bf Crowdsourcing and Research with Human Subjects}
    \item[] Question: For crowdsourcing experiments and research with human subjects, does the paper include the full text of instructions given to participants and screenshots, if applicable, as well as details about compensation (if any)? 
    \item[] Answer: \answerNA{} % Replace by \answerYes{}, \answerNo{}, or \answerNA{}.
    \item[] Justification: the paper does not involve crowdsourcing nor research with human subjects.
    \item[] Guidelines:
    \begin{itemize}
        \item The answer NA means that the paper does not involve crowdsourcing nor research with human subjects.
        \item Including this information in the supplemental material is fine, but if the main contribution of the paper involves human subjects, then as much detail as possible should be included in the main paper. 
        \item According to the NeurIPS Code of Ethics, workers involved in data collection, curation, or other labor should be paid at least the minimum wage in the country of the data collector. 
    \end{itemize}

\item {\bf Institutional Review Board (IRB) Approvals or Equivalent for Research with Human Subjects}
    \item[] Question: Does the paper describe potential risks incurred by study participants, whether such risks were disclosed to the subjects, and whether Institutional Review Board (IRB) approvals (or an equivalent approval/review based on the requirements of your country or institution) were obtained?
    \item[] Answer: \answerNA{} % Replace by \answerYes{}, \answerNo{}, or \answerNA{}.
    \item[] Justification: the paper does not involve crowdsourcing nor research with human subjects
    \item[] Guidelines:
    \begin{itemize}
        \item The answer NA means that the paper does not involve crowdsourcing nor research with human subjects.
        \item Depending on the country in which research is conducted, IRB approval (or equivalent) may be required for any human subjects research. If you obtained IRB approval, you should clearly state this in the paper. 
        \item We recognize that the procedures for this may vary significantly between institutions and locations, and we expect authors to adhere to the NeurIPS Code of Ethics and the guidelines for their institution. 
        \item For initial submissions, do not include any information that would break anonymity (if applicable), such as the institution conducting the review.
    \end{itemize}

\end{enumerate}

\end{document}